\documentclass[conference]{IEEEtran}
\IEEEoverridecommandlockouts
\usepackage{cite}
\usepackage{amsmath,amssymb,amsfonts}
\usepackage{amsthm}
\usepackage{algorithm}
\usepackage{algpseudocode}
\usepackage{graphicx}
\usepackage{textcomp}
\usepackage{xcolor}
\usepackage{enumerate} 
\usepackage{subcaption} 
\usepackage[colorlinks, citecolor=blue, linkcolor=blue]{hyperref}
\usepackage{float}

\theoremstyle{definition}
\newtheorem{definition}{Definition}
\newtheorem{lemma}{Lemma}
\newtheorem{Proposition}{Proposition}

\def\BibTeX{{\rm B\kern-.05em{\sc i\kern-.025em b}\kern-.08em
    T\kern-.1667em\lower.7ex\hbox{E}\kern-.125emX}}

\begin{document}

\title{
Steps Adaptive Decay DPSGD:\\
Enhancing Performance on Imbalanced Datasets with Differential Privacy with HAM10000
}

\author{
\IEEEauthorblockN{1\textsuperscript{st} Xiaobo Huang}
\IEEEauthorblockA{\textit{Guangdong Provincial Key Laboratory of IRADS}\\
\textit{Beijing Normal-Hong Kong Baptist University}\\
Zhuhai 519087, China \\
t330026062@mail.uic.edu.cn}
\and
\IEEEauthorblockN{2\textsuperscript{nd} Fang Xie}
\IEEEauthorblockA{\textit{Guangdong Provincial Key Laboratory of IRADS}\\
\textit{Beijing Normal-Hong Kong Baptist University}\\
Zhuhai 519087, China \\
fangxie@uic.edu.cn}
}

\maketitle

\begin{abstract}
When applying machine learning to medical image classification, data leakage is a critical issue. Previous methods, such as adding noise to gradients for differential privacy, work well on large datasets like MNIST and CIFAR-100, but fail on small, imbalanced medical datasets like HAM10000. This is because the imbalanced distribution causes gradients from minority classes to be clipped and lose crucial information, while majority classes dominate. This leads the model to fall into suboptimal solutions early. To address this, we propose SAD-DPSGD, which uses a linear decaying mechanism for noise and clipping thresholds. By allocating more privacy budget and using higher clipping thresholds in the initial training phases, the model avoids suboptimal solutions and enhances performance. Experiments show that SAD-DPSGD outperforms Auto-DPSGD on HAM10000, improving accuracy by 2.15\% under \(\epsilon = 3.0\) ,  \(\delta = 10^{-3}\).
\end{abstract}


\section{Introduction}
Deep learning has played an important role in current life, including in the medical, industrial, and health care areas. Unfortunately, data in the reality is always considered to be imbalanced, such as patient images and medical images used to train deep learning algorithms in the health care area, which are considered imbalanced most of the time.

Previous studies have already demonstrated the various possibilities of conducting attacks to extract private information from DL models, which indicates that the information used to train models in the health care and medical areas also has the potential to be exposed to unauthorized individuals..

More importantly, the fairness and privacy of the model face serious challenges when the training dataset is imbalanced \cite{17}. The work conducted by \cite{17} suggested that the bias in the dataset would lead to bias in the algorithm and model being trained.

Previous studies that implement differential privacy on imbalanced images dataset applied DPSGD on HAM10000 dataset \cite{32,33}.

To tackle the problem of implementing differential privacy on the imbalanced and small dataset, contribution has been done From the perspective of the data, previous study that implemented differential privacy on imbalanced images dataset applied DPSGD on HAM10000 dataset \cite{32,33}. And \cite{25} proposed that applying data pre-processing and class-weighted deep learning might help dealing with this problem under the condition of differential privacy. 

The existing work has limitations in multiple aspects. From the algorithm perspective, Previous studies find out that appling DPSGD to unbalanced datasets could only get a poor performance under small privacy budgets \cite{32}. At the same time, current improvements have mainly focused on data pre-processing and class-weighted deep learning \cite{25}, but no methods have addressed the unbalanced problem directly through algorithm design. 

To overcome these limitations, Adaptive DPSGD algorithms come to our sight. Since one straightforward approach to handling uneven misclassification costs is dynamically allocating the privacy budget during training \cite{25}. Auto-DPSGD-L \cite{29} and Auto-DPSGD-S \cite{8} are effective adaptive algorithms that is being proposed recently. Auto-DPSGD is an algorithm that adjusts the noise multiplier and clipping threshold after each epoch and each sample respectively using tCDP. Among all its variations, the variation steps’ decay performed the best. However, methods like Auto-DPSGD-S \cite{8} have not discussed the settings of steps and their decay hyperparameters. Additionally, the uniform steps arrangements proposed by it do not match the training processes of unbalanced datasets.

Thus, we introduce SAD-DPSGD, a novel approach that dynamically adjusts the noise multiplier and clipping threshold after each training step. Unlike traditional methods that use uniform steps, SAD-DPSGD arranges the training steps to be differential. This mechanism, in conjunction with Rényi Differential Privacy (RDP), leverages Proposition 1 to accommodate these modifications and accurately estimate the total privacy budget.

To investigate the effectiveness of various algorithms on the imbalanced dataset using HAM10000 dataset \cite{33} as an example, we execute different algorithm-based model training processes and study the influence of different parameters.

In conclusion, existing works have limitations when implemented on imbalanced dataset such as uniform steps setting, and automatic clipping threshold \cite{8} that is considered to be too large when it is adopted on unbalanced dataset such as HAM10000 dataset \cite{33}. To address these limitations, we design SAD-DPSGD, which is mainly composed of decaying steps clipping threshold estimation and decaying steps noise multiplier estimation. Our method uses noise multiplier and clipping threshold which decay by steps which are also linear decaying. This approach enables the use of a smaller noise multiplier in early training steps to avoid suboptimal solutions and ensures that the clipping threshold remains reasonable. Our work introduces the following contributions:
\begin{enumerate}[•]
\item We implement different algorithms on the imbalanced datasets HAM10000 \cite{33}, including DPSGD \cite{1}, Auto-DPSGD-L, Auto-DPSGD-S \cite{8}, SAD-DPSGD. And compare their effectiveness.
\item We propose a new adaptive noise multiplier decay mechanism for the imbalanced datasets using HAM10000 \cite{33} as an example. This mechanism extends the existing step decay approach from \cite{25} by increasing the noise multiplier after every few steps, thereby improving model accuracy. Our SAD-DPSGD includes both adaptive noise multiplier decay and adaptive clipping threshold decay mechanisms.
\item We investigate the impact of various hyperparameters, including decay parameters and step settings, on the accuracy and privacy of DP models using the HAM10000 dataset \cite{33}. We compare the performance of different algorithms (DPSGD-TS, DPSGD, Auto-DP-S, Auto-DP-L) on this unbalanced dataset.
\item Through extensive evaluation, we show that SAD-DPSGD outperforms other methods on the HAM10000 dataset \cite{33}. Specifically, it achieves 1.00\%, 0.99\%, and 0.80\% higher accuracy than Auto-DPSGD under different privacy budgets (\(\epsilon = 3\), \(\epsilon = 8\), \(\epsilon = 16\)) with \(\delta = 10^{-3}\).
\end{enumerate}

\section{RELATED WORK}
\subsection{Adaptive DPSGD Algorithm}
Many algorithms improve on DPSGD based on the idea of adaptiveness. Developed over time, \cite{2} proposed an adaptive learning rate based on DPSGD.\cite{10} developed a dynamic DPSGD method that dynamically adjusts the clipping and noise multipliers based on prior work. The DPSGD Global algorithm raised by \cite{30} and its variation DPSGD Global-Adapt proposed by \cite{22} are popular adaptive gradient scaling algorithms for DPSGD. \cite{8} also proposed an  adaptive clipping algorithm that conducts Experiments validates on the NSL-KDD dataset. \textbackslash{}cite\{21\} also proposed a novel method, DPAdaMod-AGC, by adaptively estimating the overall clipping threshold to improve the accuracy of the classification model and achieved a 2.3\% improvement in classification accuracy. \cite{31} proposed an adaptive algorithm with linear noise multiplier decay per epoch. Finally, \cite{8} extended this to exponential, step, and time decay. Step decay is most efficient, applying dynamic noise multipliers and estimating clipping thresholds from current gradient norms. However, uniform step sizes are suboptimal since model fitting grows exponentially. At the same time, estimating gradient norms is challenging due to extremely large gradients. \\
\noindent These adaptive studies are all designed for general purposes, thus we aim to develop a method for imbalanced datasets.

\subsection{Implementation of Differential Privacy on Imbalanced Datasets}
Previous studies addressing differential privacy under class imbalance have primarily focused on using preprocessing methods and class-weighted deep learning techniques to enhance model performance \cite{25}. Another approach proposed leveraging local differential privacy to improve performance \textbackslash{}cite\{19\}. However, implementing DPSGD on the imbalanced HAM10000 dataset \cite{33} yielded poor results \cite{32}, with Matthews correlation coefficients on the test set reaching only 15.60\%, 37.48\%, and 42.83\% for $\epsilon$ values of 1, 8, and 20, respectively. A study by \cite{17} found that DPSGD performed significantly better on the majority class than on the minority class in highly imbalanced datasets. Without differential privacy, the accuracy gap between female (majority) and male (minority) subgroups remained stable. However, under differentially private training, this gap widened as training progressed, especially with lower privacy levels $(\epsilon=4.98, \epsilon=16.2)$, which showed more pronounced divergence in accuracy \cite{17}. Similar findings were reported by \cite{16}, who demonstrated that DPSGD training led to a more significant drop in accuracy for minority subgroups in neural networks. These observations were further supported by experiments on gender classification models, which showed lower accuracy for black faces compared to white faces on the Flickr-based Diversity in Faces (DiF) and UTKFace datasets after applying DP \cite{23, 18}. Similar disparities were observed in sentiment analysis of African-American English tweets \cite{26} and species classification on the iNaturalist dataset \cite{20}.

\noindent The above studies have raised the problems to be resolved, but didn't propose solutions to improve the performance from the perspective of algorithms. Thus, motivated by Auto-DPSGD \textbackslash{}cite\{8\}, we try to develop an algorithm to address the limitations of the prior work. 

\section{PRELIMINARIES}
\subsection{Differential Privacy}
Differential privacy is a methodology to protect sensitive information from illegal queries (Dwork, 2008).  

\begin{definition}[Differential Privacy \cite{11}]
A randomized mechanism \( F \) provides \((\epsilon,\delta)\)-Differential Privacy (DP) if for any two datasets \( D \) and \( D' \) that differ in only a single data point \( d \), \(\forall S \subseteq \text{Range}(D) \),
\begin{equation}
\Pr(F(D) \in S) < e^\epsilon \times \Pr(F(D') \in S) + \delta. \tag{1}\label{eq1}
\end{equation}
Usually, to implement randomness, we add Gaussian noise \(\mathcal{N}(0,\sigma^2 \cdot s_f^2)\) on a mechanism \( f \) with \( l_2 \) sensitivity \( s_f \) (introduced later).
\end{definition}

\begin{definition}[\( l_k \)-Sensitivity \cite{14}] 
For a function \( f : \mathcal{X}^n \to \mathbb{R}^d \), we define its \( l_k \) norm sensitivity (denoted as \( \Delta_k f \)) over all neighboring datasets \( x, x' \in \mathcal{X}^n \) differing in a single sample as:
\begin{equation}
\sup_{x, x' \in \mathcal{X}^n} \| f(x) - f(x') \|_k \leq \Delta_k f. \tag{2}\label{eq2}
\end{equation}
In this paper, we are using \(l_2\) sensitivity.
\end{definition}

\subsection{Rényi Differential Privacy}
Rényi differential privacy (RDP) is defined through Rényi divergence:  

\begin{definition}[Rényi Divergence \cite{28}] 
Given two probability distributions \( A \) and \( A' \), the Rényi divergence of order \( \alpha > 1 \) is:
\begin{equation}
D_\alpha(A \| A') = \frac{1}{\alpha - 1} \ln \mathbb{E}_{x \sim A'} \left[ \left( \frac{A(x)}{A'(x)} \right)^\alpha \right], \tag{3}\label{eq3}
\end{equation}
where \( \mathbb{E}_{x \sim A'} \) denotes the expected value of \( x \) for distribution \( A' \), and \( A(x), A'(x) \) denote densities at \( x \).
\end{definition}

\begin{definition}[Rényi Differential Privacy (RDP) \cite{24}]
For any neighboring datasets \( x, x' \in \mathcal{X}^n \), a randomized mechanism \( \mathcal{M} : \mathcal{X}^n \rightarrow \mathbb{R}^d \) satisfies \( (\alpha, R) \)-RDP if
\begin{equation}
D_\alpha(\mathcal{M}(x) \| \mathcal{M}(x')) \leq R. \tag{4}\label{eq4}
\end{equation}
\end{definition}

\begin{definition}[RDP of Gaussian Mechanism \cite{24}]
Assuming \( f \) is a real-valued function with sensitivity \( s_f \), the Gaussian mechanism for approximating \( f' \) is
\begin{equation}
f'(D) = f(D) + N \left( 0, s_f^2 \sigma^2 \right), \tag{5}\label{eq5}
\end{equation}
where \( N(0, \mu^2 \sigma^2) \) is a normally distributed random variable (standard deviation \( \mu \sigma \), mean 0). The mechanism satisfies \( (\alpha, \alpha / 2\sigma^2) \)-RDP.
\end{definition}

\begin{lemma}[Conversion from RDP to DP \cite{24}] 
If a randomized mechanism \( f : D \rightarrow \mathbb{R} \) satisfies \( (\alpha, R) \)-RDP, then it satisfies
\begin{equation}
\left( R + \ln\left(\frac{\alpha - 1}{\alpha}\right) - \frac{\ln \delta + \ln \alpha}{\alpha - 1}, \delta \right)\text{-DP} \tag{6}\label{eq6}
\end{equation}
for any \( 0 < \delta < 1 \).
\end{lemma}
\begin{Proposition}\cite{24}
Let \( f : \mathcal{D} \rightarrow \mathcal{R}_1 \) be \( (\alpha, \epsilon_1) \)-RDP and \( g : \mathcal{R}_1 \times \mathcal{D} \rightarrow \mathcal{R}_2 \) be \( (\alpha, \epsilon_2) \)-RDP, then the mechanism defined as \( (X, Y) \), where \( X \leftarrow f(D) \) and \( Y \leftarrow g(X, D) \), satisfies \( (\alpha, \epsilon_1 + \epsilon_2) \)-RDP.\\\\
Proposition 1 provides a way to estimate the privacy budget when the noise multiplier varies during training. Our algorithm’s proof is derived from Proposition 1 and the definitions above.
\end{Proposition}
\section{METHODOLOGY}
\subsection{Noise and Multiplier Decay Function}
We perform a noise and multiplier estimation to calculates noise and clipping threshold for every steps \(D_i\). \(sigma\_clip\_estimation\) is the function that is used to conduct estimation, the details is introduced as follow.
\begin{algorithm}[htbp]
\caption{$sigma\_clip\_estimation$}
\begin{algorithmic}[1]
\Require{
    overall training epochs T,
    current epoch t,
    step decay parameter $\gamma$, 
    $\sigma$ decay parameter $\beta$,
    clipping threshold decay parameter $a$,
    Final noise multiplier $\sigma_n$,
    Final clipping threshold $C_n$
}
\Ensure{
    \textbf{current noise multiplier and clipping threshold $\sigma_t,$ $C_t$}
}
\State Divide T into multiple steps \(D_0 \dots D_n\), where \(\sum_{i}^{n} D_i = T\) and \(D_i\cdot\gamma = D_{i-1}\)
\For {\(i \) in \( \{0,1,\dots,n-1\} \)}
    \State \(\sigma_i = \sigma_{i+1}\cdot\beta\) 
    \State \(C_i = C_{i+1}\cdot a\)
\EndFor
\If {t in \(D_i\)}
    \State \(\sigma_t = \sigma_{i}\) 
    \State \(C_t = C_{i}\)
\EndIf
\State \Return \(\sigma_t,C_t\)
\end{algorithmic}
\end{algorithm}
\noindent In the overall training process, the clipping threshold, of every steps \(D_i\) is considered to decaying as the training moves on, whereas the noise multiplier is increasing linearly by exponential steps. The relation ship between \(C_t\) and \(C_{t+1}\),  and \(\sigma_t\) and \(\sigma_{t+1}\) are considered to be \(C_t = C_{t+1}\cdot a\) and \(\sigma_t = \sigma_{t+1}\cdot\beta\) respectively, where $i$ is considered to be the index of the steps.

\subsection{SAD-DPSGD}
This section explains the step-adaptive decay DPSGD algorithm (SAD-DPSGD), which uses adaptive decay mechanisms for the noise multiplier and clipping threshold. The algorithm takes inputs such as the training dataset, loss function, model, decay parameters, and other settings. It initializes the model with pre-trained weights and runs for \( T \) iterations. In each iteration, it calculates the clipping threshold and noise multiplier using the \( \text{sigma\_clip\_estimation} \) function, computes gradients, performs gradient clipping, adds noise, and updates the model with the modified gradients. The sampling probability is \( q = B/|trainingDataset| \), where \( B \) is the batch size.\\
\indent The motivation of using an exponential step setting, which divides the total iterations \textit{T} into multiple steps \(D_0, \dots, D_n\), where \(\sum_{i}^{n} D_i = T\) and \(D_i\cdot \gamma = D_{i-1}\), stems from observations of the model training process. Figure \hyperref[fig1]{1} demonstrates that the accuracy of the minority group rises dramatically at the very beginning but then gradually slows down as the iterations increase under DPSGD conditions. In contrast, the majority group has already achieved a very high accuracy and maintains a reasonably high accuracy throughout the subsequent training process. The non-linear growth of the accuracy of the minority group indicates that the division of the steps shouldn't be considered to be uniform.\\
\indent  The core of the SAD-DPSGD algorithm lies in its noise and clipping threshold mechanisms. Unlike Auto-DPSGD, which starts with the maximum noise multiplier, SAD-DPSGD begins with a low noise level to avoid sub-optimal solutions in the early training stages. Meanwhile, the clipping threshold starts high and gradually decays as training progresses, aligning with the typical training process. These initial settings help SAD-DPSGD outperform other variants, especially on imbalanced datasets.\\
\indent Algorithm 2 is compatible with DP. The whole computations use per-sample gradients only and clip it before updates. We demonstrate the estimation of the total privacy budget should be as follow:\\
\indent For every step with \(\sigma = \sigma_t\) and  clipping threshold equal to \(C_t\), thus the sensitivity is considered to be \(C_t\). Undered these condition, for step t the rdp is considered to be \[rdp_t =  D_\alpha(\mathcal{M}(x) \| \mathcal{M}(x')),\] where \[M'(D) = M(D) + N \left( 0, C_t^2 \sigma_t^2 \right)\] according to \hyperref[eq1]{(1)}, \hyperref[eq2]{(2)}, \hyperref[eq3]{(3)}, \hyperref[eq4]{(4)} and \hyperref[eq5]{(5)}. \\
The overall $rdp$ is \[rdp=\sum_{t=0}^{n}rdp_t,\] 
where $n$ is the number of steps, $t$ indicates the index of current step, $\alpha$ is the order of the Rényi divergence, The parameter $\delta$ represents the probability that plain $\epsilon$-differential privacy could be violated. Then we can convert it from RDP to DP with Lemma \hyperref[eq6]{1}:
  \[ \left( rdp + \ln\left(\frac{\alpha - 1}{\alpha}\right) - \frac{\ln \delta + \ln \alpha}{\alpha - 1}, \delta \right)\text{-DP}\]

\begin{algorithm}[H]
\caption{SAD-DPSGD}
\begin{algorithmic}[1]
\Require{
    training datasets $\{x_1, x_2, \dots, x_N\}$, 
    loss function $\mathcal{L}(\theta, x_i)$. 
    Parameters: 
    learning rate $\alpha$, 
    Batch size for training $B$, 
    Final noise multiplier $\sigma_{n}$, 
    Final clipping bound for training $C_{n}$, 
    clipping decay parameter $\beta$,
    noise multiplier decay parameter $\gamma$, 
    number of steps n,
}
\Ensure{
    \textbf{the final trained model $w_t$}
}
\While{$t < T$}
    \State Randomly sample a batch \(\mathcal{B}_{t}\) with $B$ batch size and with probability \(\frac{|\mathcal{B_t}|}{N}\);
    
    \State \(\sigma_t,C_t = sigma\_clip\_estimation(T,t,\gamma,\beta,a,\sigma_n,C_n)
    \)
    
    \For{$x_i \in \mathcal{B}_t$}
        \State Compute $g_t(x_i) \gets \nabla \mathcal{L}(w_t, x_i)$
        \State $\bar{g}_t(x_i) \gets g_t(x_i) / \max\left(1, \frac{\|g_t(x_i)\|_2}{C_j}\right)$
    \EndFor
    \State $\widetilde{\bar{g}}_t \gets \frac{1}{|\mathcal{B}_t|} \left( \sum_{x_i \in B_t} \bar{g}_t(x_i) + \mathcal{N}(0, \sigma_t^2 C_t^2) \right)$
    \State $w_{\text{t}} = w_{t-1} - \alpha \widetilde{\bar{g}}_t$
\EndWhile
\State \Return $w_T$
\end{algorithmic}
\end{algorithm}

\begin{figure}[htbp] 
    \vspace{-10pt}
    \centering
    \begin{subfigure}{\linewidth}
        \centering
        \includegraphics[width=1\linewidth]{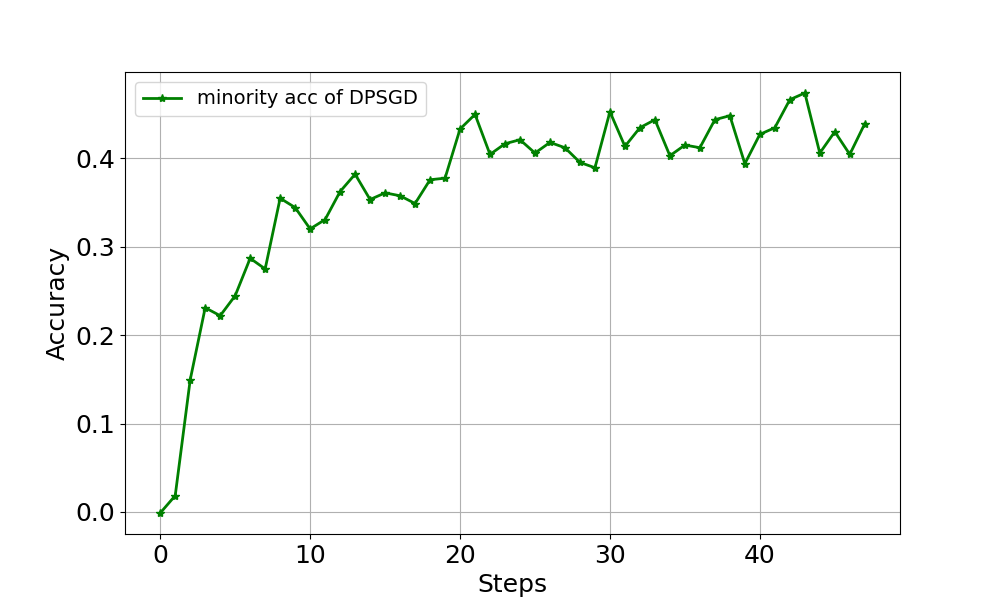}
        \caption{minority group accuracy of DPSGD}
    \end{subfigure}%
    \vfill
    \begin{subfigure}{\linewidth}
        \centering
         \includegraphics[width=1\linewidth]{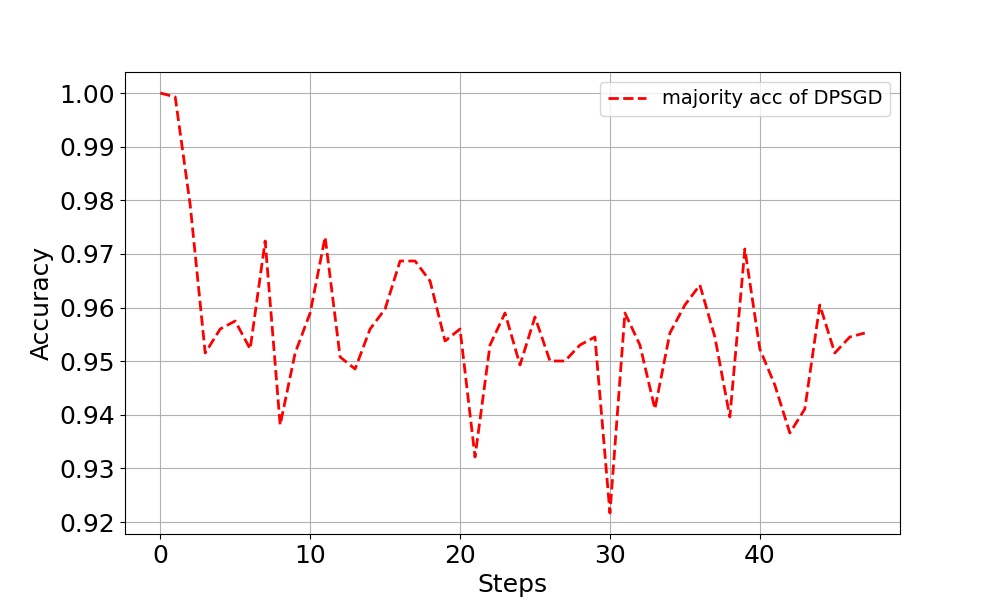}
        \caption{majority group accuracy of DPSGD}
    \end{subfigure}%
    \caption{the performance of the model during the training}
    \label{fig1}
\label{fig1}
\end{figure}
\section{EXPERIMENTS}
\begin{figure}[htbp] 
    \vspace{-10pt}
    \centering
    \begin{subfigure}{\linewidth}
         \centering
        \includegraphics[width=\linewidth]{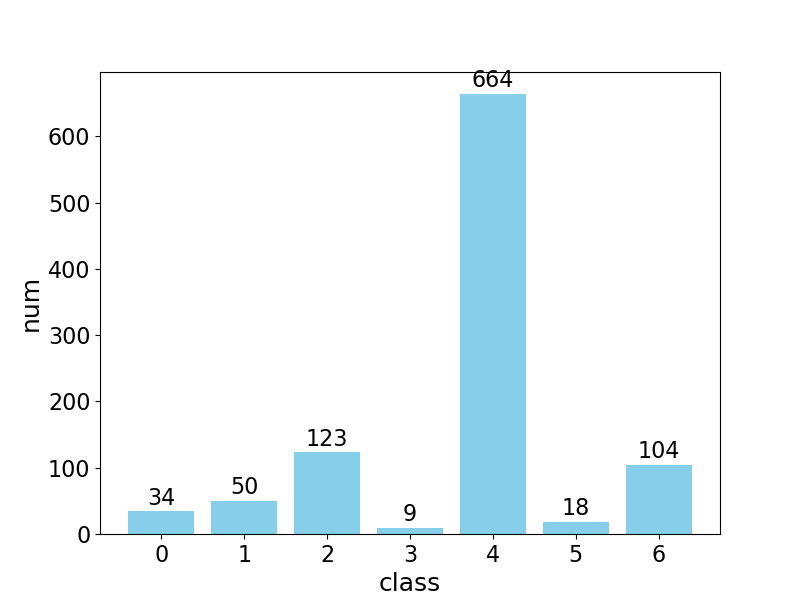}
        \captionsetup{font=large, labelfont=large} % 
        \caption{distribution of test data}
    \end{subfigure}%
    \vfill
    \begin{subfigure}{\linewidth}
        \centering
        \includegraphics[width=\linewidth]{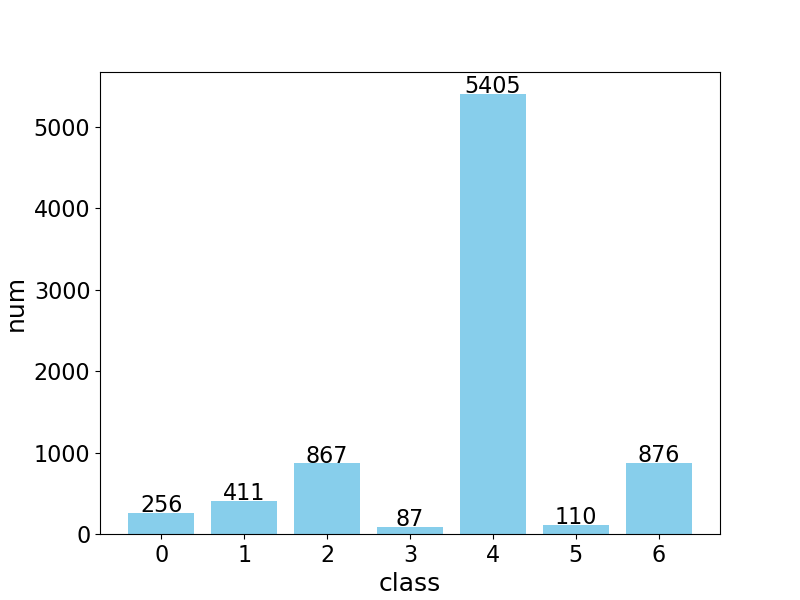}
        \captionsetup{font=large, labelfont=large} % 
        \caption{distribution of train data}
    \end{subfigure}%
    \caption{distribution of HAM10000 dataset}
\label{fig2}
\end{figure}

Without loss of generality, we use HAM10000 \cite{33} to verify the effectiveness of the different algorithms. Without losing fairness, we chose to use RDP account for all the algorithm to test the effectiveness of the setting of different decay parameters. The implementation is available on \href{https://github.com/FangXieLab/SAD-DPSGD}{GitHub}. HAM10000 \cite{33} is a imbalanced dataset dominated by one major class: Melanocytic Nevi (NV),and consists of 10,015 dermatoscopic images categorized into seven classes labeled from 0 to 6, including Actinic keratoses (AK), Basal cell carcinoma (BCC), Benign keratosis-like lesions (BKL), Dermatofibroma (DF), Melanocytic nevi (NV), Vascular lesions (VASC), and Melanoma (MEL). The distribution of the overall dataset is presented as follows in Figure \hyperref[fig2]{2}.
\subsection{Effect of Parameters on Noise Multipliers' Distribution}
The performance of the model depends on the variances, means, and initial values of the noise multiplier, as well as the length of the initial step. By analyzing these factors, we can identify a reasonable range for the parameters to verify further. The mean of the noise multipliers is calculated as: \(mu = \left(\frac{1}{n}\right) \sum_{i=1}^{n} s_i \cdot \sigma_i \), where \( n \) is the number of steps, \( s_i \) is the length of each step, and \( \sigma_i \) is the noise multiplier for each step. A lower mean suggests better performance. The variance is calculated as: \( \text{var} = \sum_{i=1}^{n} s_i \cdot (\sigma_i - \mu)^2 \). A larger variance indicates that the noise multiplier in the final step could be very large, which is undesirable. The initial noise multiplier is the value used in the first step, and we aim to keep it as small as possible. We first investigate how these variances, means, and initial values change as the parameters vary.
\subsubsection{Effect of decay parameter of noise multiplier (\(\beta\)) on noise multipliers distribution}
\begin{figure*}[htbp]
    \vspace{-10pt}
    \centering
    \begin{subfigure}[b]{0.33\linewidth}
        \centering
        \includegraphics[width=\linewidth]{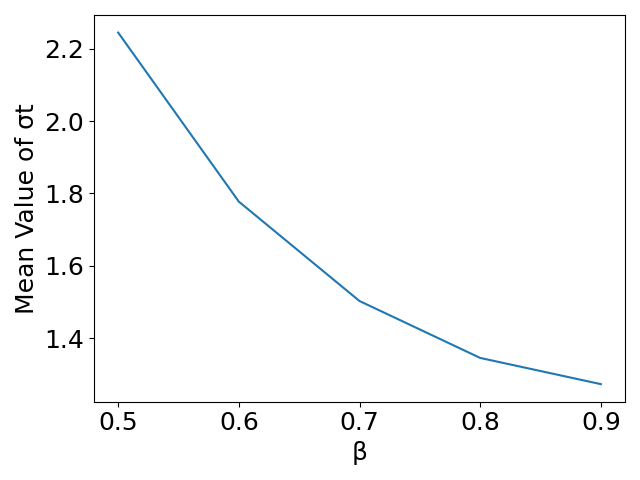}
        \captionsetup{font=small, labelfont=small} % Set caption and label font size to small
    \end{subfigure}%
    \begin{subfigure}[b]{0.33\linewidth}
        \centering
        \includegraphics[width=\linewidth]{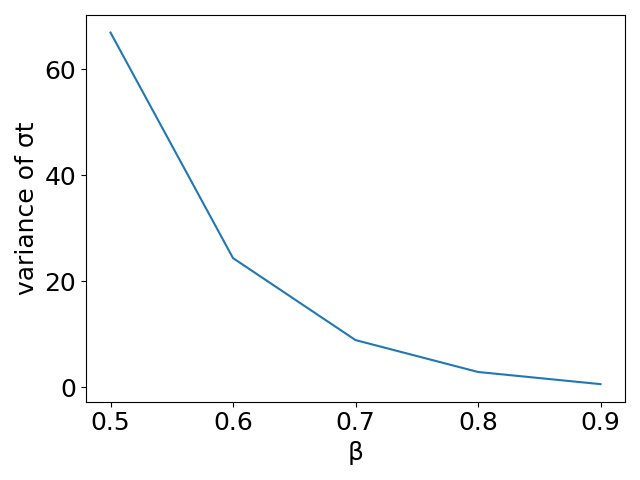}
        \captionsetup{font=small, labelfont=small} % Set caption and label font size to small
    \end{subfigure}%
    \begin{subfigure}[b]{0.33\linewidth}
        \centering
        \includegraphics[width=\linewidth]{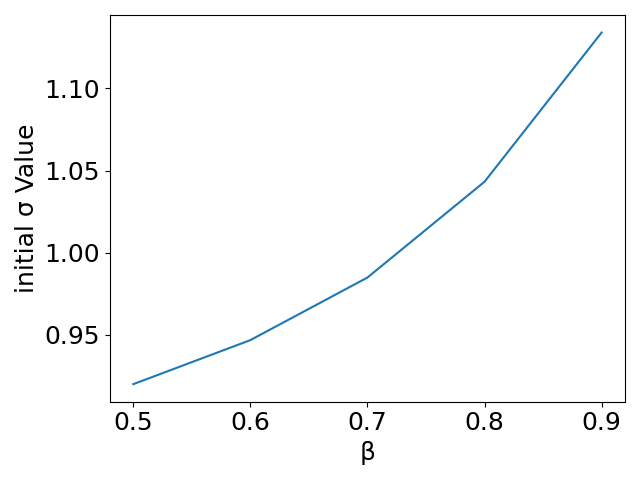}
        \captionsetup{font=small, labelfont=small} % Set caption and label font size to small
    \end{subfigure}%
\label{fig3}
    \caption{relationship between \(\sigma_t\) and $\beta$}
\end{figure*}

In order to primarily investigate the effect of $\beta$ on the distribution of $\sigma$, we fixed $\gamma$ to be $0.9$ and $n=3$. The below Figure \hyperref[fig3]{3} demonstrates the results. The tendency of the change is quite straightforward. The mean value and variance of the $\sigma$ decrease as the $\beta$ increase whereas the initial $\sigma$ increase as $\beta$ increases. Since we wanted the initial $\sigma$ to be as small as possible and mean value of the $\sigma$ to be as small as possible, we need to find a balance between mean, variance and initial $\sigma$. Thus, a reasonable range of $\sigma$ could be from $0.6$ to $0.8$.

\subsubsection{Effect of Decay Parameter of Steps ($\gamma$) on Noise Multipliers Distribution}
\begin{figure*}[htbp]
    \vspace{-10pt}
    \centering
    \begin{subfigure}[b]{0.24\linewidth}
        \centering
        \includegraphics[width=\linewidth]{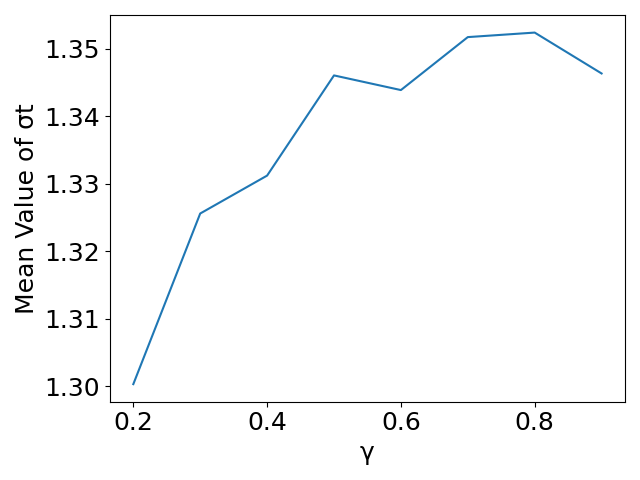}
        \captionsetup{font=small, labelfont=small}
        \caption{}
    \end{subfigure}%
    \hfill
    \begin{subfigure}[b]{0.24\linewidth}
        \centering
        \includegraphics[width=\linewidth]{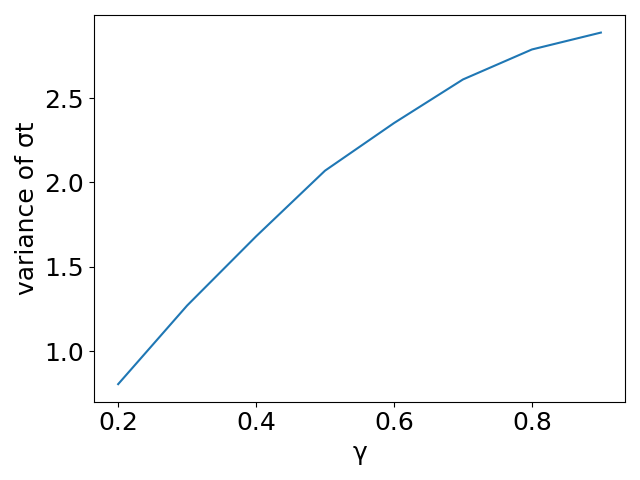}
        \captionsetup{font=small, labelfont=small}
        \caption{}
    \end{subfigure}%
    \hfill
    \begin{subfigure}[b]{0.24\linewidth}
        \centering
        \includegraphics[width=\linewidth]{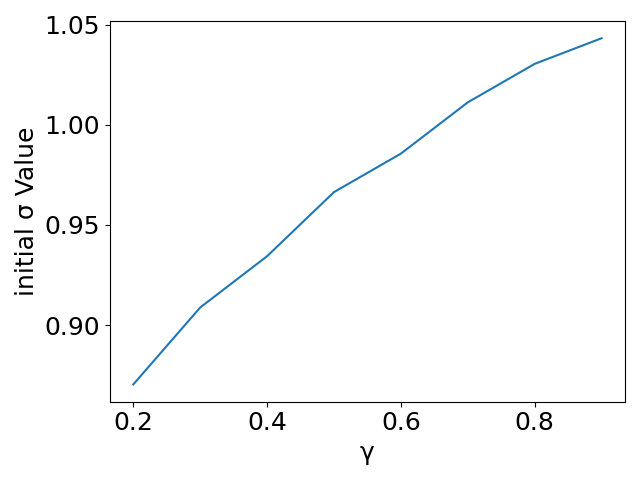}
        \captionsetup{font=small, labelfont=small}
        \caption{}
    \end{subfigure}%
    \hfill
    \begin{subfigure}[b]{0.24\linewidth}
        \centering
        \includegraphics[width=\linewidth]{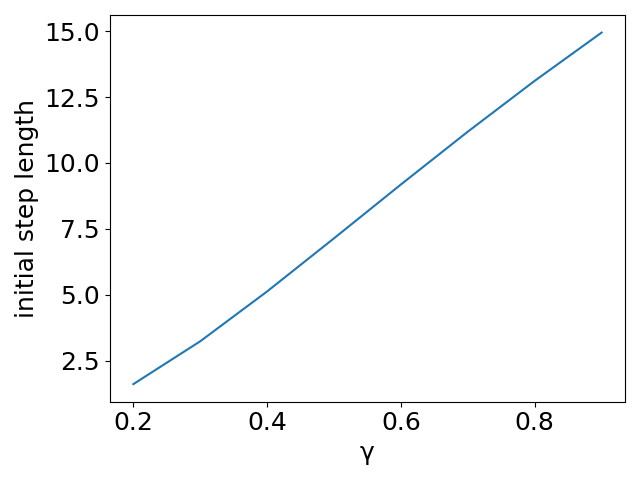}
        \captionsetup{font=small, labelfont=small}
        \caption{}
    \end{subfigure}
    \caption{Relationship between \(\sigma_t\) and \(\gamma\)}
    \label{fig4}
\end{figure*}

To understand how $\gamma$ affects the distribution of $\sigma$, we set n to 3 and $\beta$ to 0.8. Figure \hyperref[fig4]{4} shows the results. As $\gamma$ decreases, the mean, variance, and initial $\sigma$ all tend to be smaller. However, we should also consider the length of the initial step. Figure \hyperref[fig4]{4}(d) indicates that the initial step length increases as $\gamma$ increases. For example, when $\gamma=0.2$, the initial step length is only 1/50 of the total steps, making it less significant compared to DPSGD. Therefore, Figure \hyperref[fig4]{4} provides limited insights for further discussion. Instead, we should verify the impact of $\gamma$ on performance by considering the initial step length and balancing the trade-offs between the initial $\sigma$ and the mean of $\sigma_t$.  

\subsubsection{Influence of the Number of the Steps (n) on Noise Multipliers Distribution}
\begin{figure*}[htbp]
    \vspace{-10pt}
    \centering
    \begin{subfigure}[b]{0.33\linewidth}
        \centering
        \includegraphics[width=\linewidth]{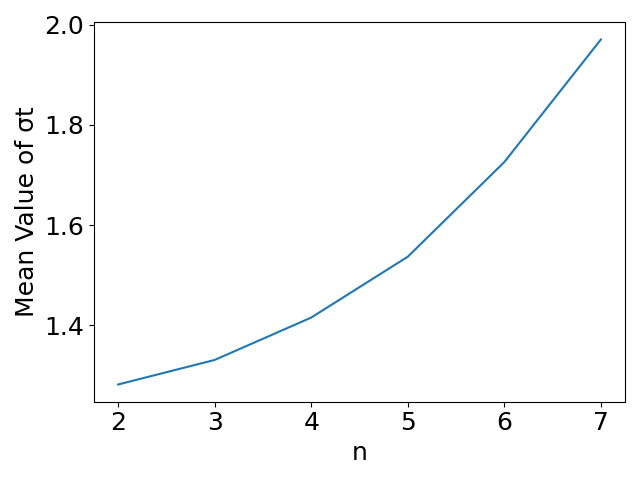}
        \captionsetup{font=small, labelfont=small} % Set caption and label font size to small
    \end{subfigure}%
    \hfill
    \begin{subfigure}[b]{0.33\linewidth}
        \centering
        \includegraphics[width=\linewidth]{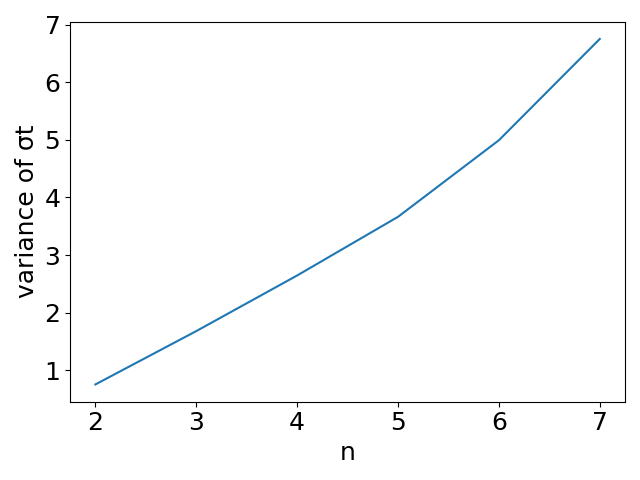}
        \captionsetup{font=small, labelfont=small} % Set caption and label font size to small
    \end{subfigure}%
    \hfill
    \begin{subfigure}[b]{0.33\linewidth}
        \centering
        \includegraphics[width=\linewidth]{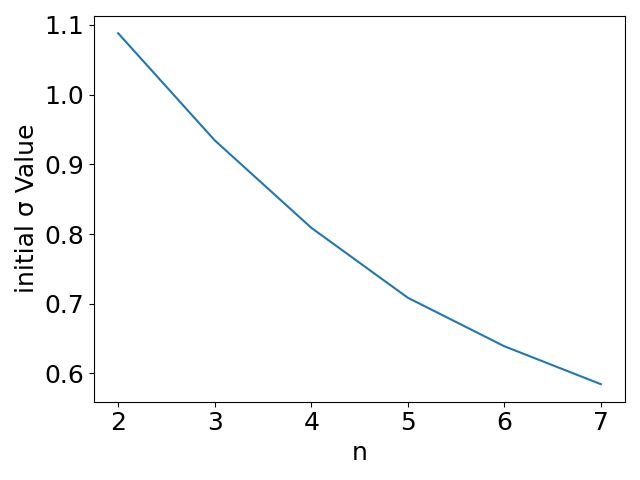}
        \captionsetup{font=small, labelfont=small} % Set caption and label font size to small
    \end{subfigure}
    \caption{relationship between \(\sigma_t\) and $n$}
\label{fig6}
\end{figure*}

To study the effect of $n$ on the distribution of $\sigma$, we fixed $\gamma$ at $0.9$ and $\beta$ at $0.9$. Figure \hyperref[fig6]{6} shows the results. As n increases, the mean and variance of $\sigma$ generally increase, while the initial $\sigma$ decreases. Since we aim to minimize both the initial $\sigma$ and the mean of $\sigma$, we need to find a balance among the mean, variance, and initial $\sigma$. Based on these observations, a reasonable range for n is between $3$ and $5$.

\subsection{Effect of Parameters on Performance}
To further validate the reasonableness of this parameter setting, we used the HAM10000 dataset \cite{33} for testing. Our goal is to suggest a possible parameter setting that maximizes the performance. 
\subsubsection{Effect of Decay Parameter of Noise Multiplier (\(\beta\)) on Performance}
\begin{figure}[htbp]
    \centering
    \includegraphics[width=0.7\linewidth]{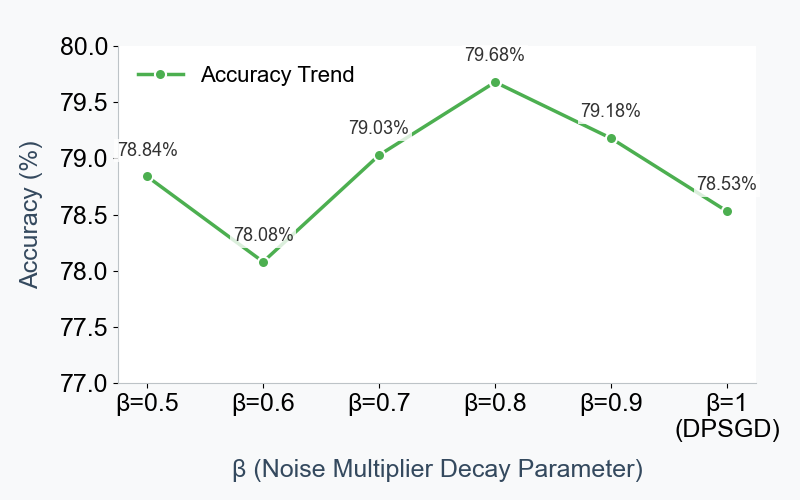}
    \caption{Effects of $\sigma$ decay parameter with $\epsilon=3.0$, $\delta=1e-3$, $\gamma=1$,  $n=3$}
    \label{fig7}
    \vspace{-10pt}
\end{figure}

To find the optimal \(\beta\), we conducted experiments with 3 steps, keeping \(n\) as a parameter for later discussion. Based on previous analysis, we tested \(\beta\) values from 0.6 to 0.9, aiming to balance the mean and variance of \(\sigma_t\). Figure \hyperref[fig7]{6} shows that model performance peaks at \(\beta = 0.8\) and declines as \(\beta\) deviates from this value. This indicates an optimal \(\beta\) setting: too high (\(\beta\) close to 1) reduces the decay effect, while too low causes excessive variance in \(\sigma_t\) values. Thus, \(\beta = 0.8\) best balances these trade-offs and maximizes performance.
\subsubsection{Effect of Decay Parameter of Steps ($\gamma$) on Performance}
\begin{figure}[htbp]
    \vspace{-9.8pt}
    \centering
    \includegraphics[width=0.7\linewidth]{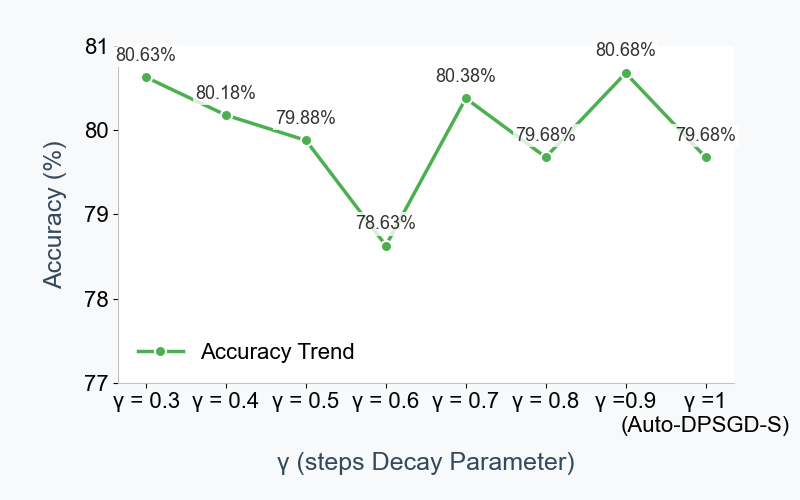}
    \caption{Effects of steps decay parameter with $\epsilon=3.0$, $\delta=1e-3$, $\beta=0.8$, $n=3$}
\label{fig8}
\vspace{-8pt}
\end{figure}

To determine the optimal value of $\gamma$, we conducted experiments while keeping the number of steps at 3 and varying $\gamma$ from 0.3 to 1. The number of steps n is also a parameter to be discussed later. Figure \hyperref[fig8]{7} illustrates how the decay parameter $\gamma$ affects model performance. Performance peaks at $\gamma$=0.9 and $\gamma$=0.3, and declines as $\gamma$ deviates from these values in either direction. This indicates that a moderately large $\gamma$ or moderately small $\gamma$ within the range $(0,1)$ is suitable for the training process. These results correspond to our previous study. When $\gamma$ is small, even if the length of the initial steps isn't large, the $\gamma$ is small enough to make a difference. When $\gamma$ is large, even if the initial $\sigma$ is not as small as when $\gamma$ is small, the larger length of the initial step compensates for the drawback. This suggests that the optimal point to balance the trade-offs between the initial $\sigma$ and the length of the initial step could be at $\gamma=0.9$ or $\gamma=0.3$.

\subsubsection{Influence of the Number of the Steps (n) on Performance}
To better understand the impact of the steps decay mechanism, we analyzed how the number of steps \( n \) affects model performance, especially in unbalanced datasets. We set \( \beta = 0.8 \) and \( \gamma = 0.9 \), and tested the model performance under the conditions \( \epsilon = 3 \) and \( \delta = 10^{-3} \) within the suggested range of \( n \in [3, 5] \). The results are shown in Table \hyperref[t3]{1}. Table \hyperref[t3]{1} shows that \( n = 3 \) is likely the optimal solution, with performance worsening as \( n \) increases or decreases. This aligns with our previous analysis that there is an optimal \( n \) within 2 to 5, balancing the trade-offs between the initial \( \sigma \) and the mean and variance of \( \sigma_t \) to achieve the best model performance.

\begin{table}[htbp]
    \centering
    \begin{tabular}{c|c|c|c}
        \hline
        \textbf{$n$ (number of steps)}& \textbf{$n=3$} & \textbf{$n=4$} & \textbf{$n=5$} \\
        \hline
        accuracy & $80.63\%$ & $79.13\%$ & $78.98\%$ \\
        \hline
    \end{tabular}
    \label{tab:steps_decay}
    \vspace{0.5em} % 增加与下方说明的间距
    \caption{Effects of number of steps with $\epsilon=3.0$, $\delta=1e-3$, $\beta=0.9$, $\gamma = 0.9$}
\label{t3}
\end{table}

\begin{figure*}[htbp]
    \vspace{-10pt}
    \centering
    \begin{subfigure}[b]{0.33\linewidth}
        \centering
        \includegraphics[width=\linewidth]{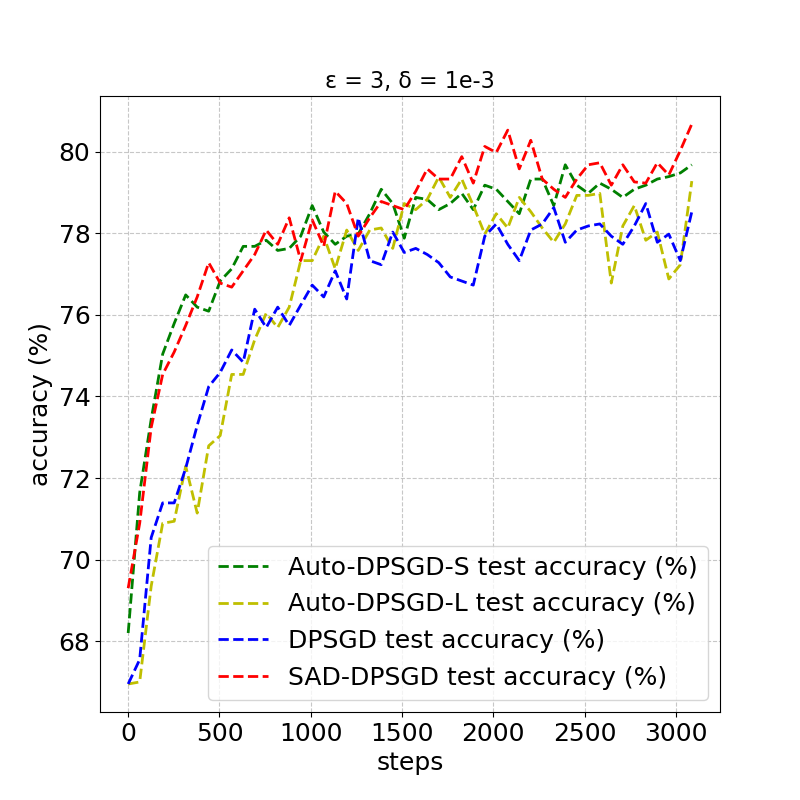}
        \captionsetup{font=large, labelfont=large} % Set caption and label font size to small
    \end{subfigure}%
    \hfill
    \begin{subfigure}[b]{0.33\linewidth}
        \centering
        \includegraphics[width=\linewidth]{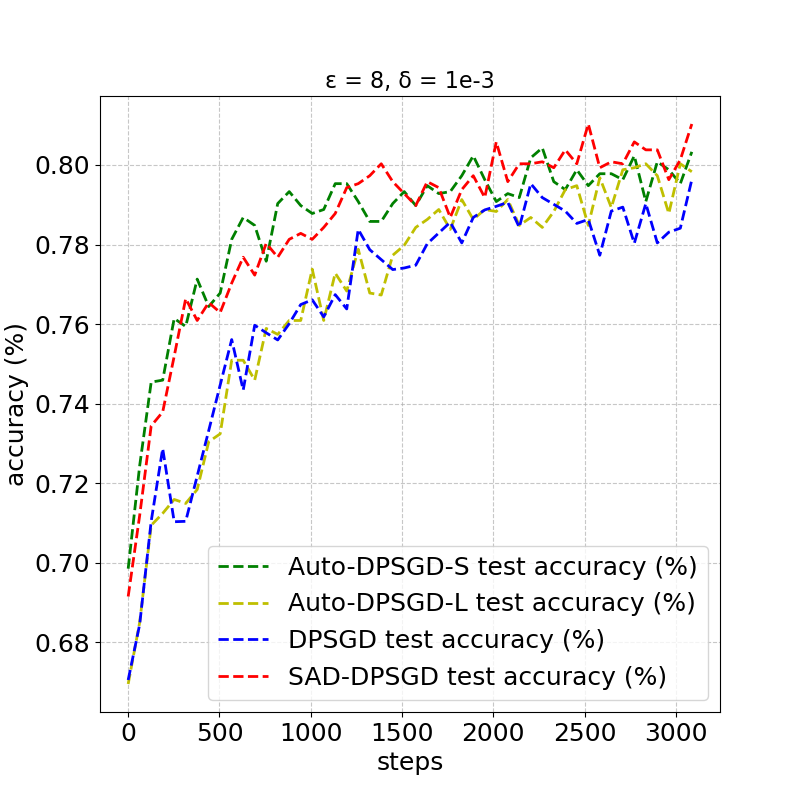}
        \captionsetup{font=large, labelfont=large} % Set caption and label font size to small
    \end{subfigure}%
    \hfill
    \begin{subfigure}[b]{0.33\linewidth}
        \centering
        \includegraphics[width=\linewidth]{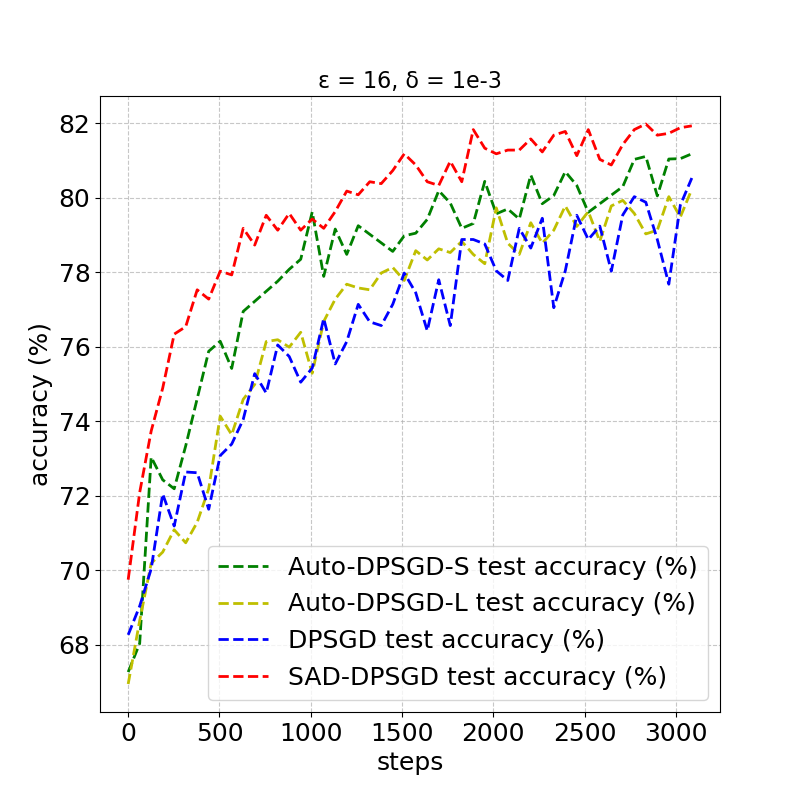}
        \captionsetup{font=large, labelfont=large} % Set caption and label font size to small
    \end{subfigure}%
    \caption{performance of different algorithms}
\label{fig9}
\end{figure*}

\begin{figure}[htbp]
    \vspace{-10pt}
    \centering
    \begin{subfigure}[b]{\linewidth}
        \centering
        \includegraphics[width=1\linewidth]{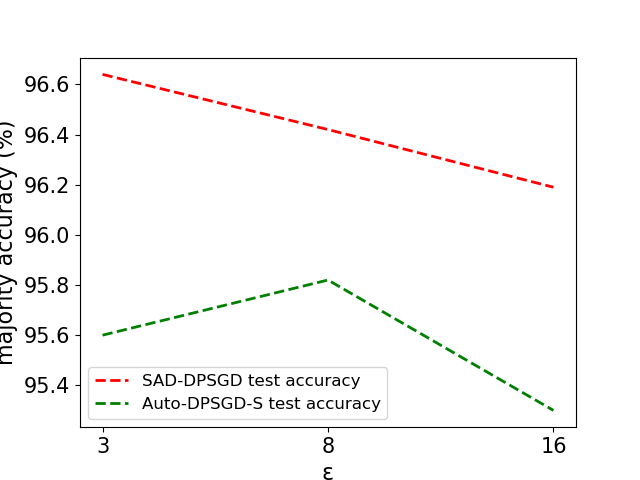}
        \captionsetup{font=large, labelfont=large} % Set caption and label font size to small
    \end{subfigure}%
    \vfill
    \begin{subfigure}[b]{\linewidth}
        \centering
        \includegraphics[width=1\linewidth]{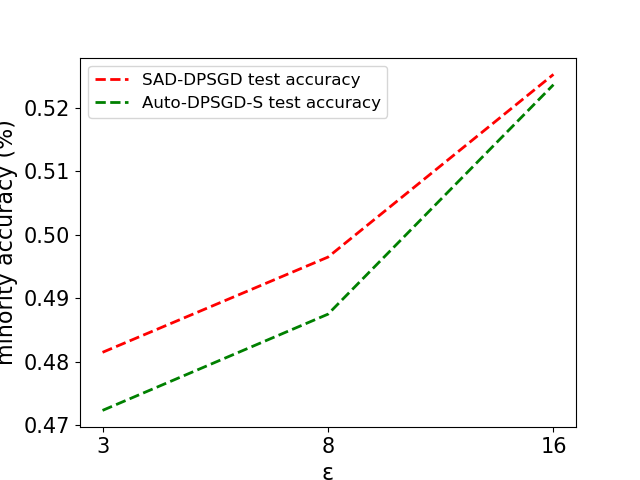}
        \captionsetup{font=large, labelfont=large} % Set caption and label font size to small
    \end{subfigure}%
    \caption{comparison of accuracy of the majority group and the minority group}
\label{fig10}
\end{figure}

\subsection{Performance of Different Algorithms within Different Privacy Setting}
Figure \hyperref[fig9]{8} and Table \hyperref[t4]{2} show the overall performance of different algorithms. We can see that SAD-DPSGD outperforms the other algorithms on the HAM10000 dataset \cite{33} on all privacy settings, indicating the effectiveness of SAD-DPSGD in the case of imbalanced dataset. And Figure \hyperref[fig10]{9} demonstrates the advantages of SAD-DPSGD in a more straightforward manner. The model's performance using SAD-DPSGD is superior for both the majority and minority groups compared to the model using Auto-DPSGD-S. These results prove that our method effectively addresses the challenges posed by imbalanced datasets to a certain extent.

\begin{table}[htbp]
    \centering
    \begin{tabular}{p{1.2cm}|p{1cm}|p{1.5cm}|p{1.5cm}|p{1.5cm}}
        \hline
        \textbf{Algorithm} & \textbf{DPSGD} & \textbf{Auto DPSGD-S} & \textbf{Auto DPSGD-L} & \textbf{SAD-DPSGD}  \\
        \hline
        $\epsilon=3$ & $78.53\%$  & \centering$79.68\%$ & \centering $79.28\%$  & $\mathbf{80.68\%}$\\
        $\epsilon=8$ & $79.63\%$  & \centering$80.33\%$ & \centering $80.03\%$  & $\mathbf{81.03\%}$\\
        $\epsilon=16$ & $80.53\%$  & \centering$81.18\%$ & \centering $80.23\%$ & $\mathbf{81.98\%}$\\
        \hline
    \end{tabular}
    \label{tab:steps_decay}
    \vspace{0.5em} % 增加与下方说明的间距
    \caption{Performance of different algorithms with $\delta=1e-3, \beta=0.8, \gamma = 0.9,n=3$}
\label{t4}
\end{table}

\section{CONCLUSIONS AND DISCUSSIONS}
We demonstrate that current adaptive DPSGD algorithms can address challenges posed by imbalanced datasets in differential privacy. However, we identify limitations in the decay mechanism of previous Adaptive DPSGD algorithms. Our new SAD-DPSGD algorithm outperforms predecessors in handling imbalanced datasets under differential privacy. While this paper verifies SAD-DPSGD's effectiveness on the HAM10000 dataset \cite{33} and provides optimal hyperparameter settings, future work could focus on testing SAD-DPSGD on other imbalanced datasets. The variability of optimal hyper parameters across datasets is a limitation of our current work.

\section*{Acknowledgement}
\noindent F. Xie was supported in part by the Guangdong Basic and Applied Basic Research Foundation (grant number 2023A1515110469), in part by the Guangdong Provincial Key Laboratory IRADS (grant number 2022B1212010006), and in part by the grant of Higher Education Enhancement Plan (2021-2025) of "Rushing to the Top, Making Up Shortcomings and Strengthening Special Features" (grant number 2022KQNCX100).

\bibliographystyle{IEEEtran}

\bibliography{references}

\end{document}